\documentclass{article}

\usepackage{arxiv}

\usepackage[utf8]{inputenc} 
\usepackage[T1]{fontenc}    
\usepackage{hyperref}       
\usepackage{url}            
\usepackage{booktabs}       
\usepackage{amsfonts}       
\usepackage{nicefrac}       
\usepackage{microtype}      
\usepackage{lipsum}		
\usepackage{graphicx}
\usepackage{natbib}
\usepackage{doi}
\usepackage{graphicx}
\usepackage{amsmath}
\usepackage{booktabs}
\usepackage[dvipsnames]{xcolor}
\usepackage{tikz}
\usepackage{graphicx}
\usepackage{mwe}
\usepackage{subcaption}
\usepackage{float}
\usepackage{adjustbox}

\title{\textbf{Confidence-Nets:}  A Step Towards better Prediction Intervals for regression Neural Networks on small datasets}

\author{ \hspace{1mm}Mohamedelmujtaba Altayeb\thanks{Corresponding Author} \\
    Key Laboratory of C \& PC Structures Ministry of Education\\
    Southeast University, Nanjing, 210096, China\\
	\texttt{mohamed.elmujtaba.o@outlook.com} \\
	\And
	\hspace{1mm}Abdelrahman M. Elamin \\
	Structural and Geotechnical Engineering Department\\
	University of Naples Federico II, Napoli, Italy\\
	\texttt{a0alsir@gmail.com} \\
    \And
	\hspace{1mm}Hozaifa Ahmed \\
    Structural and Geotechnical Engineering Department\\
	University of Naples Federico II, Napoli, Italy\\
	\texttt{hozzaifaa@gmail.com} \\
    \And
	\hspace{1mm}Eithar Elfatih Elfadil Ibrahim \\
	Department of Statistics \& Computer Science\\
    Faculty of Mathematical Sciences and Informatics\\
    University of Khartoum, Khartoum, Sudan\\
	\texttt{Etharalfath75@gmail.com} \\
    \And
	\hspace{1mm}Omer Haydar\\
	Department of Electrical and Electronic Engineering\\
    University of Khartoum, Khartoum, Sudan\\
	\texttt{omer.2002.hayder@gmail.com}\\
    \And
	Saba Abdulaziz\\
	Department of Electrical and Electronic Engineering\\
    University of Khartoum, Khartoum, Sudan\\
	\texttt{sabaabdulbasit24@gmail.com} \\
    \And
	Najlaa H. M. Mohamed\\
	Department of Electrical and Electronic Engineering\\
    University of Khartoum, Khartoum, Sudan\\
	\texttt{najlaahassabelnabi@gmail.com} \\
}

\renewcommand{\shorttitle}{\textit{Confidence-Nets}}

\hypersetup{
pdftitle={Confidence-Nets: A Step Towards better Prediction Intervals for regression Neural Networks on small datasets},
pdfsubject={Confidence-Nets},
pdfauthor={Mohamedelmujtaba Altayeb , Abdelrahman M. Elamin , Hozaifa Ahmed , Either E. E. Ibrahim , Omer Haydar , Saba Abdulaziz , Najlaa H. M. Mohamed},
pdfkeywords={Deep Neural Networks, XGBoost , Ensemble Learning, Prediction Intervals, Regression},
}

\begin{document}
\maketitle

\begin{abstract}
	The recent decade has seen an enormous rise in the popularity of deep learning and neural networks. These algorithms have broken many previous records and achieved remarkable results. Their outstanding performance has significantly sped up the progress of AI, and so far various milestones have been achieved earlier than expected. However, in the case of relatively small datasets, the performance of Deep Neural Networks (DNN) may suffer from reduced accuracy compared to other Machine Learning models. Furthermore, it is difficult to construct prediction intervals or evaluate the uncertainty of predictions when dealing with regression tasks. In this paper, we propose an ensemble method that attempts to estimate the uncertainty of predictions, increase their accuracy and provide an interval for the expected variation. Compared with traditional DNNs that only provide a prediction, our proposed method can output a prediction interval by combining DNNs, extreme gradient boosting (XGBoost) and dissimilarity computation techniques. Albeit the simple design, this approach significantly increases accuracy on small datasets and does not introduce much complexity to the architecture of the neural network. The proposed method is tested on various datasets, and a significant improvement in the performance of the neural network model is seen. The model’s prediction interval can include the ground truth value at an average rate of 71\% and 78\% across training sizes of 90\% and 55\%, respectively. Finally, we highlight other aspects and applications of the approach in experimental error estimation, and the application of transfer learning. 
\end{abstract}


\section{1.	Introduction}
Deep neural networks are the algorithms at the core of the deep learning field. These algorithms have allowed us to make significant advances in the field of AI, from achieving super-human level performance on atari games and beating humans at the game of GO (\cite{Mnih2015,Geron2017}) to autonomous driving (\cite{Wang2020}) and protein folding (\cite{Jumper2021}). While many other machine learning algorithms focus on learning only one or two layers of representation, deep neural networks can automatically learn tens or hundreds of successive layers of representations from exposure to training data (\cite{Chollet2018}). These representations are primarily learned through the hidden layers of neural networks, which are usually stacked on top of each other in the sense that each layer's output is the input of the next layer. Adding more layers stacked on top of each other allows the neural network to learn more representations, providing more accurate predictions. \\
A large number of scientists are devoted to the study of interpreting models that are considered black-box models (\cite{Fan2021}). A small portion of this research focuses on uncertainty quantification (UQ) and Prediction Intervals (PI) construction. This is an incredibly challenging task, as reviewed in studies such as (\cite{Tsuji2021}), who discussed the difficulty of error estimation of deep learning models. (\cite{Minakowski2023}) developed a posteriori error estimator for neural network approximations of partial differential equations. (\cite{Lai2022}) Explored the uncertainty in regression neural networks and then proposed a prediction interval construction method from the statistics perspective. A large-scale evaluation of empirical frequentist coverage properties was also provided by (\cite{Kompa2021}). Their evaluation focused on well-known uncertainty quantification techniques on a suite of regression and classification tasks. \\
Furthermore, (\cite{Hirschfeld2020}) examined several methods on five regression datasets and concluded that existing UQ methods are inadequate for all common-use cases and further research is needed. A recently published study by (\cite{Abdar2021}) reviewed the recent advances in uncertainty quantification methods used in deep learning, highlighting the fundamental research challenges and directions associated with UQ. Most of these studies eventually demonstrate the difficulty of uncertainty quantification and building prediction intervals in regression neural networks. Despite the various attempts at uncertainty quantification, the current methods are challenging to implement. Each method introduces more steps to the already complicated nature of how neural networks arrive at a prediction.\\
Moreover, introducing more complexity into the deep neural network can affect its structure and thus hinder the application of transfer learning, an essential feature of neural networks. In cases such as in extensive experimentation, transfer learning plays an essential role as it can be applied to scenarios of experimental conditions change, such as changes in Spatio-temporal variables, changes in material characteristics, or the presence of a consistent source of error unknown to the practitioner (un-noticeable minor machine faults). This further applies to fields such as the study of construction materials, where small datasets belonging to the cementitious composite family are usually available and share many similarities. Even in the case of the development of systems that can fully explain the inner workings of neural network prediction, there will remain the question of how this understanding can help improve accuracy. \\
In the meantime, we believe it's convenient to sacrifice the complete understanding of deep neural networks and settle for the estimation of error as an approach to uncertainty quantification and constructing prediction intervals. The ability to estimate the prediction errors can allow us to improve accuracy by directly correcting errors, as well as maintain some of the neural network's core features, such as transfer learning. In this paper, we aim to develop a simple ensemble method that indirectly combines an artificial neural network with an extreme gradient boosted trees (XGBoost) algorithm and dissimilarity equations for improving the predictions of a neural network. Previous research has shown that ensemble techniques that combine neural networks and tree-based methods are highly effective with small datasets (\cite{Altayeb2021}). The proposed method allows the improvement of prediction accuracy without disturbing the main dynamics of the neural network, therefore allowing the efficient scaling and application of transfer learning. Hereafter, we will refer to our method as “confidence-nets”. \\
In section 2 of the paper, we present the datasets, algorithms, and the combination method, followed by the training process for the ensemble approach. The paper focuses mainly on presenting the application of error correction to improve accuracy, estimation, and the method's robustness. Section 3 introduces error correction results and shows how the ensemble method is much more effective than a similar neural network model for small datasets.\\

\section{Methods}
\subsection{Data collection}
The datasets used in this work and their corresponding details are specified in the table (\ref{tab:datasets_used_in_study}). Big data and massive datasets are rare in some fields, such as the study of construction and building materials, mainly concrete. This may be due to the considerable cost of experiments to collect such datasets. For these reasons, we thought that construction and building materials datasets represent a good test bed for our proposed concept. We will mainly consider small datasets found online for the approach proposed in this paper. All datasets are normalized using min-max normalization, then split at different train: test ratios to evaluate the model's performance compared with the base neural network model. \\

\begin{table}[H]
	\caption{Datasets used in this study.}
	\centering
	\begin{tabular}{p{0.03\textwidth}p{0.5\textwidth}p{0.1\textwidth}p{0.1\textwidth}p{0.1\textwidth}}
		\toprule
		\# & Dataset name & Publishing date	& \# Features (columns)	& \# Instances (rows) \\
		\midrule

		1 & Concrete Slump Test (\cite{Yeh2007}) & 2009-04-30 & 10 & 103 \\
		2 & Concrete Compressive Strength (\cite{Yeh2007}) & 2007-08-03 & 9 & 1030 \\
		3 & Dataset of 622 Rectangular Concrete-filled steel tube columns (\cite{NguyenThiMai2021}) & 2021-06-08 & 7 & 623 \\
		4 & Strain Hardening Cementitious Composites Data for FDNN ensemble predictive model (\cite{AltayebMohamedelmujtaba;WangXin;Musa2021}) & 2021-03-03 & 29 & 241 \\
		5 & Compressive strength dataset of foamed\/normal concrete for prediction model (\cite{NGUYENTUAN;NGO2018}) & 2018-11-20& 9 & 1133 \\
		6 & Compressive strength of alkali-activated binder concrete cured at ambient temperature (\cite{RamagiriKruthiKiran;BoindalaSrimanPankaj;Zaid2021}) & 2021-03-03 & 11 & 372 \\
		7 & A Dataset of Compressive Strength of FRCM-confined Concrete Columns (\cite{IrandeganiMohammadAli;ZhangDaxu;Shadabfar2021}) & 2021-08-17 & 5 & 76 \\
		8 & Fake transfer length data generated using generative adversarial networks (\cite{JeongHoseong;HanSun-Jin;ChoiSeung-Ho;KimJaeHyun;Kim2020}) & 2020-08-25 & 5 & 10002 \\
		9 & Experimental dataset (\cite{ChenHuaicheng;QianChunxiang;LiangChengyao;Kang}) & 2018-01-18 & 10 & 639 \\
		\bottomrule
	\end{tabular}
	\label{tab:datasets_used_in_study}
\end{table}

\subsection{Model development}
\begin{figure}[H]
	\centerline{\includegraphics[width=6in, height=5.8in]{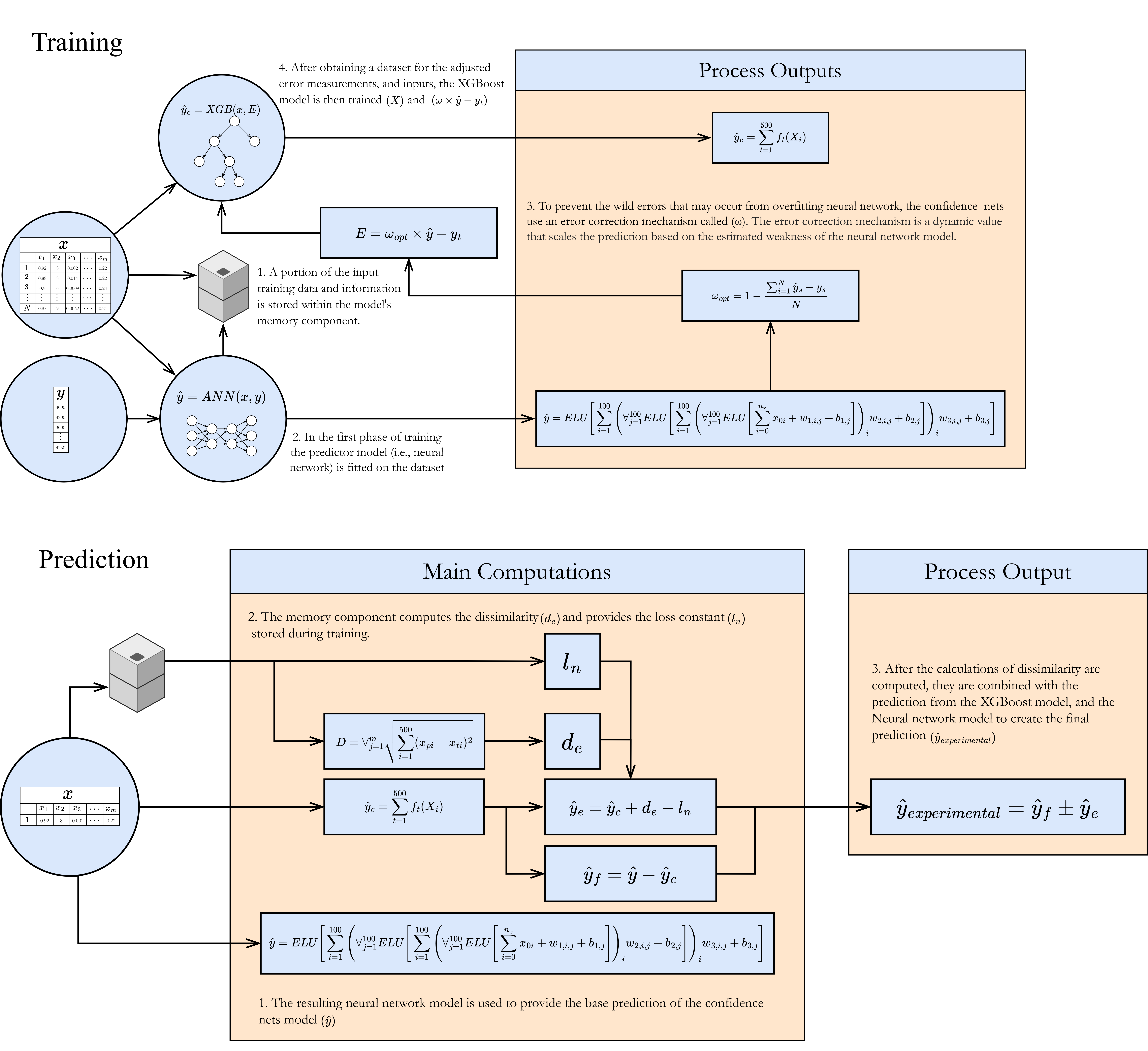}}
	\caption{Illustration of the training procedure and prediction process of the confidence nets model.}
	\label{fig:illustration_of_confidence_nets_algorithm}
\end{figure}

The confidence nets model comprises three components, the primary prediction model (a neural network model), 
an error estimation model (XGBoost model), and a memory component. 
The memory component is responsible for storing some information about training, 
which is used in later stages to estimate standard errors and build the prediction 
interval. 
Both the neural network model and error estimation model have the same inputs; 
however, the output of the neural network is the prediction ($\hat{y}$) 
The target parameter and the output of the error estimation model are an estimation of the expected variations ($\hat{y}_e$) 
between the prediction and the actual experimental result. 
This means that the neural network model is responsible for predicting the outputs. 
However, the error estimation model is responsible 
for predicting the difference between the output and the actual experimentally obtained values. 
The expected variation ($\hat{y}_e$) then becomes an indicator of how confident the neural network thinks that it will get the prediction right, 
hence the term \"confidence\" in the name of the model, while the \"net\" stands for the base neural network. 
By sharing the same inputs as the neural network model, the error estimation model must learn the relationship between a set of inputs and the error produced, 
subsequently understanding the weakness of the neural network model. 
The two models work synchronously to provide a prediction in the form ($\hat{y} \pm \hat{y}_e$)\; 
therefore, the final prediction and an indication of experimental success are given 
through an informative prediction of the confidence nets model. 
Fig. \ref{fig:illustration_of_confidence_nets_algorithm} illustrates the training and prediction process of the confidence nets model. \\

The main difference in the approach is in the training of the model. The training of the model takes two stages. In the first training phase, the neural network model (i.e., neural network) is usually fitted on the dataset. The confidence nets model implemented in this study comprises three layers, a convolution layer for extracting information and a hidden layer for improving the modelling capabilities of the model. Here it is important to note that this would be considered a complex model for a small task and would result in overfitting. Still, as will be shown later, the confidence nets architecture takes several measures to address this issue, for one the model uses an error correction mechanism that corrects the model's predictions when they are off by a large margin, as will be explained in more detail in the following sections. \\
If implemented correctly, a convolutional layer can significantly boost the feature extraction capabilities of a deep neural network model. Instead of the typical 2D convolution algorithm used for image-based tasks, we use a 1D convolution operation. The convolutional layers receive the input ($X^m$) as a tensor of channels ($K_m$) and compute a new output ($X^{m+1}$) composed of ($O_m$) channels. The output of each channel is called a feature map and can be computed as \\

\begin{equation}
	X^{(m)}_o=\sum_{k}^{n}W^{(m)}_{ok}*X^{(m-1)}_k+b^{(m)}_o
\end{equation}

Where (*) denotes the (1D) convolutional operation. The matrix ($W^{(m)}_{ok}$) parameterizes a spatial filter that the layer can use to detect or enhance some features in the incoming input. The specific action of this filter is automatically learnt from data during the model's training. \\
For training the neural network model, 
the Adam optimizer is used. 
The Huber loss is the neural network's primary loss function. 
Although the most common loss function used 
for most regression tasks (such as this task) is the mean squared error, 
Huber loss provides more flexibility to outliers, as seen in equation (\ref{eq:huber}). 
This equation essentially says that for loss values less than $\delta$, 
the mean squared error is used; for loss values greater than $\delta$, 
the mean absolute error is used. 
This effectively combines the best properties of the two loss functions. 
The equations of the three losses are given below.

\begin{equation}\label{eq:huber}
	L(y,\hat{y})_{huber}=
	\begin{cases}
	\frac{1}{2}(y - \hat{y})^2 & for \left | y - \hat{y} \right | \leq \delta\\
	\delta \bullet (\left | y - \hat{y} \right | - \frac{1}{2}\delta ) & for \left | y - \hat{y} \right | >  \delta \\
	\end{cases}       
\end{equation}

\begin{equation}
	L(y,\hat{y})_{mse}=\frac{1}{n} \sum_{i=1}^{n}(y - \hat{y})^2
\end{equation}

\begin{equation}
	L(y,\hat{y})_{mae}=\frac{1}{n} \sum_{i=1}^{n}\left | y - \hat{y} \right |
\end{equation}
\\
A portion of the input training data is stored within the model's memory without including any information about the output targets. This memory is used later to compute a similarity score between a new input and previously trained inputs. Some wild errors may occur from overfitting the neural network; thus, the confidence nets use an error correction mechanism. The error correction mechanism is a dynamic value that is added to the prediction (similar to the bias term) based on the estimated weakness of the neural network model.  \\
Although machine learning, in general, can model the relationships in data with satisfying performance. In actual experiments, obtaining the same values predicted by machine learning or a deep learning model can be complicated. The success of experimental trials relies upon several factors that are difficult to account for. However, our proposed method makes several precautions to provide safer predictions. After the first phase, once a trained neural network is obtained, it is used to create the error estimation dataset. The error estimation dataset comprises the original inputs and the error between the prediction and the obtained experimental result. This measured error is then slightly scaled up/down by multiplying the prediction with the parameter $\omega$. The factor ($\omega$) can be obtained by averaging a sample of the predictions and experimental values to determine if the network tends to mostly overestimate or underestimate its predictions. In case of overestimation, it forces the correction of the predictions to be lower and vice versa. This ensures a more inclusive error estimation property, a safer error margin, and less dependence on the neural network if its predictions are wildly off. After obtaining a dataset for the adjusted error measurements, and inputs, the XGBoost model is then trained using the inputs and targets in $X_e$. \\

\begin{equation}\label{eq:omega_opt}
	\omega_{opt}=1-\frac{\sum_{i=1}^{N}(\hat{y} - y)}{N} 
\end{equation}

\begin{equation}
	E=\omega*\hat{y}-y
\end{equation}

\begin{equation}
	X_e=[X;E]
\end{equation}
\\
Although multiplication with $\omega$ results in an artificially induced measure of safety for the predictions, it is necessary to produce a reliable performance, as will be seen later. The parameter $\omega$ is calculated automatically by the confidence nets model during training using equation (\ref{eq:omega_opt}). \\
Then an error estimation dataset composed of the errors of predictions is created ($X_e$) using the trained neural network models predictions. Next, the neural network model is combined with the error estimation model to give the final prediction ($\hat{y_f}$). The final prediction is provided by subtracting the estimated error from the original prediction, as shown in equation (\ref{eq:y_hat_f}). \\

\begin{equation}
	h_1 = ELU(\forall^{100}_{j=1} \sum_{i=1}^{n_x} X_{O_i}w_{1,i,j} + b_{1,j})
\end{equation}

\begin{equation}
	h_2 = ELU(\forall^{100}_{j=1} \sum_{i=1}^{100} h_{1,i}w_{2,i,j} + b_{2,j})
\end{equation}

\begin{equation}
	\hat{y} = ELU(\forall^{100}_{j=1} \sum_{i=1}^{100} h_{2,i}w_{3,i,j} + b_{3,j})\\
\end{equation}

\begin{equation}
	\hat{y_c} = XGBoost(X_e) = \sum_{i=1}^{500} f_t(X_{e_i})\\
\end{equation}

\begin{equation}\label{eq:y_hat_f}
	\hat{y_f} = \hat{y} - \hat{y_c}\\
\end{equation}

The second major component that allows for dynamic estimations of error is the dissimilarity factor ($d_e$). This factor is based on the Euclidean distance equation and is a computation of the amount of difference between the inputs of the current sample and the samples in the memory. Since a small amount of data is usually used to train a confidence nets model, The model's memory component can store this training data. The dissimilarity factor establishes a relationship between a new input and the trained data and computes a normalized dissimilarity value, as shown in equation (\ref{eq:dissimilarity_factor}). This factor provides an additional measure of safety in the estimation of error. If a sample is closely related to samples in the memory, then this factor approaches zero; when the sample is very strange to the confidence nets memory, this factor increases, increasing the estimated error. Finally, the average loss during training is stored in the model's memory as a constant $l_n$ and is used to adjust the bias in experimental error estimation. 

\begin{equation}\label{eq:dissimilarity_factor}
	D = d(x_p,x_t)=\forall^{N}_{j=1} \sqrt{\sum_{i=1}^{n_x} (x_{p_i} - x_{t_i})}\\
\end{equation}

\begin{equation}
	d_e = min([D])\\
\end{equation}

\begin{equation}
	\hat{y_e} = \hat{y_c} + d_e - l_n\\
\end{equation}

\begin{equation}
	\hat{y_e} = \hat{y_c} + d_e - l_n\\
\end{equation}

\begin{equation}
	\hat{y}_{experimental} = \hat{y_f} \pm \hat{y_e}\\
\end{equation}

To evaluate the ability to capture the estimate of errors using the prediction interval ($\hat{y}_{experimental}$), the results of the predictions of the confidence nets model are compared to a base neural network model on the datasets introduced previously in table \ref{tab:datasets_used_in_study}. Throughout this study, the hyperparameters used were kept constant for both models. For comparison, both models were built using Pytorch (\cite{PaszkeAdamandGrossSamandMassaFranciscoandLererAdamandBradburyJamesandChananGregoryandKilleenTrevorandLinZemingandGimelsheinNataliaandAntigaLucaandDesmaisonAlbanandKopfAndreasandYangEdwardandDeVitoZacha2019}), and for the error estimation model, the extreme boosting gradient from XGBoost was used (\cite{Chen2016}).

\section{Results}
\subsection{Error prediction}
\begin{figure}[H]
	\centering
	\begin{subfigure}{0.45\textwidth}
		\includegraphics[width=\textwidth]{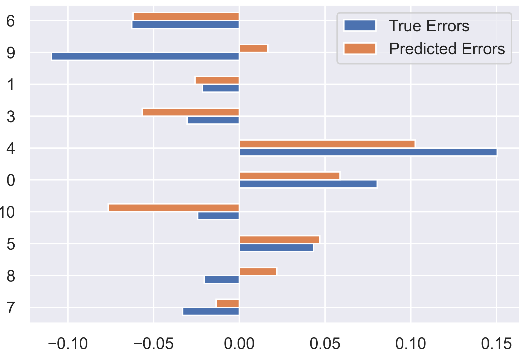}
		\caption{Training Size (90\%)}
	\end{subfigure}
	\hfill
	\begin{subfigure}{0.45\textwidth}
		\includegraphics[width=\textwidth]{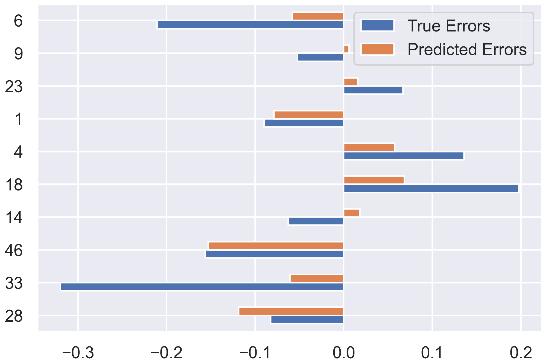}
		\caption{Training Size (55\%)}
	\end{subfigure}
	\hfill
	\caption{Dataset: Concrete Slump Test}
	\label{fig:14}
\end{figure}

\begin{figure}[H]
	\centering
	\begin{subfigure}{0.45\textwidth}
		\includegraphics[width=\textwidth]{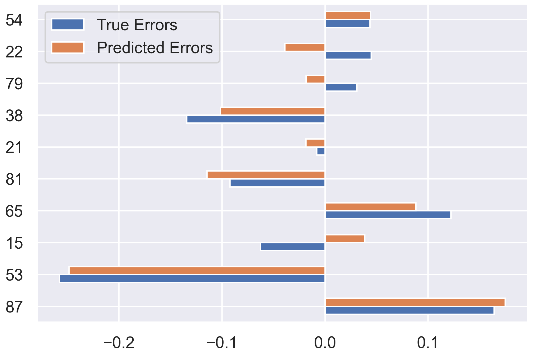}
		\caption{Training Size (90\%)}
	\end{subfigure}
	\hfill
	\begin{subfigure}{0.45\textwidth}
		\includegraphics[width=\textwidth]{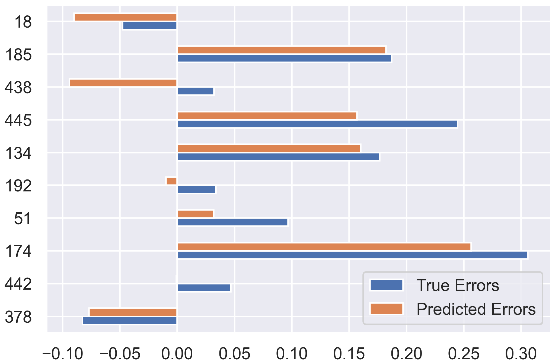}
		\caption{Training Size (55\%)}
	\end{subfigure}
	\hfill
	\caption{Dataset: Concrete Compressive Strength}
	\label{fig:15}
\end{figure}

\begin{figure}[H]
	\centering
	\begin{subfigure}{0.45\textwidth}
		\includegraphics[width=\textwidth]{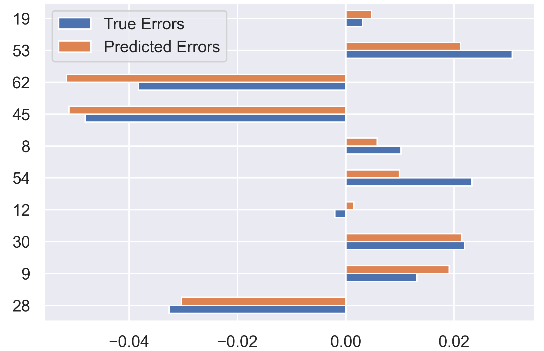}
		\caption{Training Size (90\%)}
	\end{subfigure}
	\hfill
	\begin{subfigure}{0.45\textwidth}
		\includegraphics[width=\textwidth]{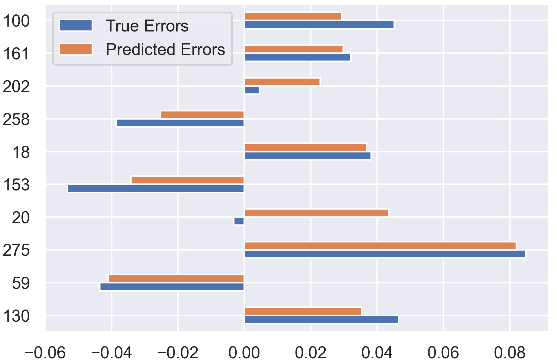}
		\caption{Training Size (55\%)}
	\end{subfigure}
	\hfill
	\caption{Dataset: 622 Rectangular Concrete-filled steel tube columns}
	\label{fig:16}
\end{figure}

\begin{figure}[H]
	\centering
	\begin{subfigure}{0.45\textwidth}
		\includegraphics[width=\textwidth]{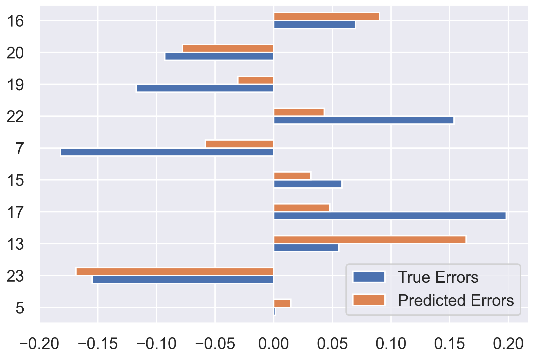}
		\caption{Training Size (90\%)}
	\end{subfigure}
	\hfill
	\begin{subfigure}{0.45\textwidth}
		\includegraphics[width=\textwidth]{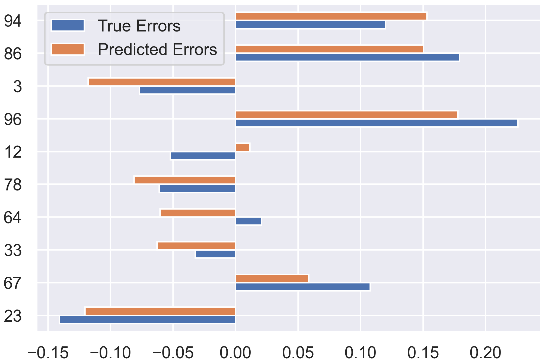}
		\caption{Training Size (55\%)}
	\end{subfigure}
	\hfill
	\caption{Dataset: Strain Hardening Cementitious Composites Data for FDNN 	ensemble predictive model}
	\label{fig:17}
\end{figure}

\begin{figure}[H]
	\centering
	\begin{subfigure}{0.45\textwidth}
		\includegraphics[width=\textwidth]{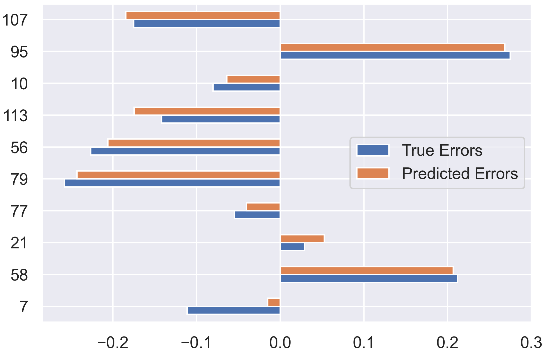}
		\caption{Training Size (90\%)}
	\end{subfigure}
	\hfill
	\begin{subfigure}{0.45\textwidth}
		\includegraphics[width=\textwidth]{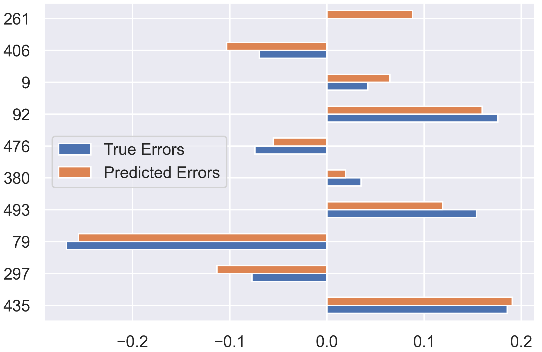}
		\caption{Training Size (55\%)}
		\label{fig:18}
	\end{subfigure}
	\hfill
	\caption{Dataset: Compressive strength dataset of foamednormal concrete for prediction model}
\end{figure}

\begin{figure}[H]
	\centering
	\begin{subfigure}{0.45\textwidth}
		\includegraphics[width=\textwidth]{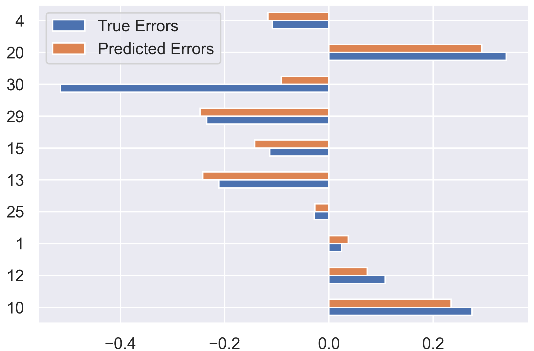}
		\caption{Training Size (90\%)}
	\end{subfigure}
	\hfill
	\begin{subfigure}{0.45\textwidth}
		\includegraphics[width=\textwidth]{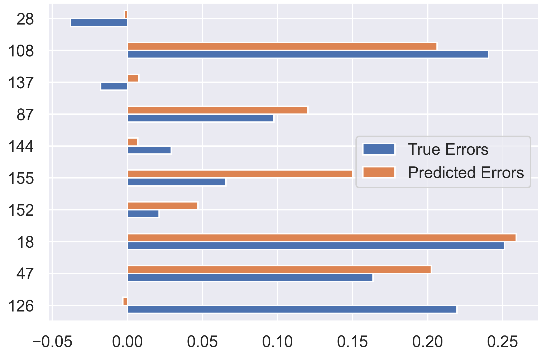}
		\caption{Training Size (55\%)}
	\end{subfigure}
	\hfill
	\caption{Dataset: Compressive strength of alkali-activated binder concrete cured at ambient temperature}
	\label{fig:19}
\end{figure}

\begin{figure}[H]
	\centering
	\begin{subfigure}{0.45\textwidth}
		\includegraphics[width=\textwidth]{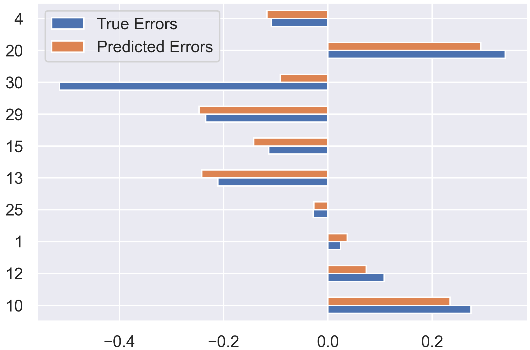}
		\caption{Training Size (90\%)}
	\end{subfigure}
	\hfill
	\begin{subfigure}{0.45\textwidth}
		\includegraphics[width=\textwidth]{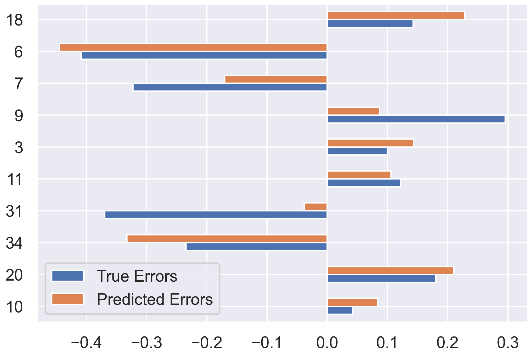}
		\caption{Training Size (55\%)}
	\end{subfigure}
	\hfill
	\caption{Dataset: A Dataset of Compressive Strength of FRCM-confined Concrete Columns}
	\label{fig:20}
\end{figure}

\begin{figure}[H]
	\centering
	\begin{subfigure}{0.45\textwidth}
		\includegraphics[width=\textwidth]{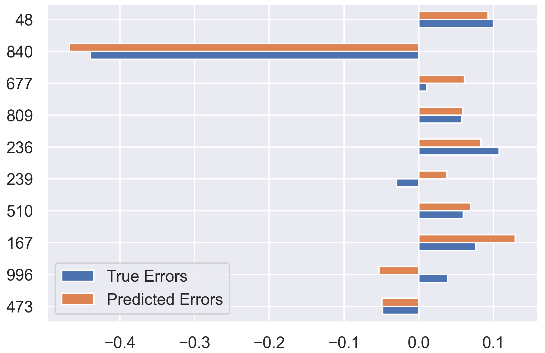}
		\caption{Training Size (90\%)}
	\end{subfigure}
	\hfill
	\begin{subfigure}{0.45\textwidth}
		\includegraphics[width=\textwidth]{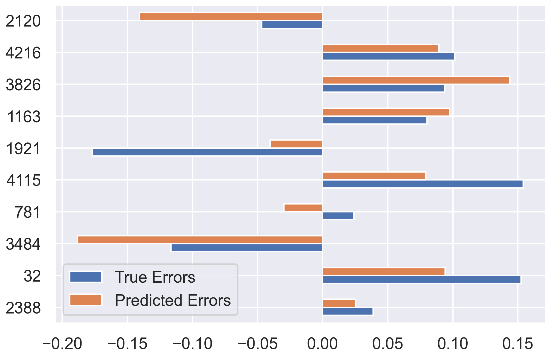}
		\caption{Training Size (55\%)}
	\end{subfigure}
	\hfill
	\caption{Dataset: Fake transfer length data generated using generative adversarial networks}
	\label{fig:21}
\end{figure}

\begin{figure}[H]
	\centering
	\begin{subfigure}{0.45\textwidth}
		\includegraphics[width=\textwidth]{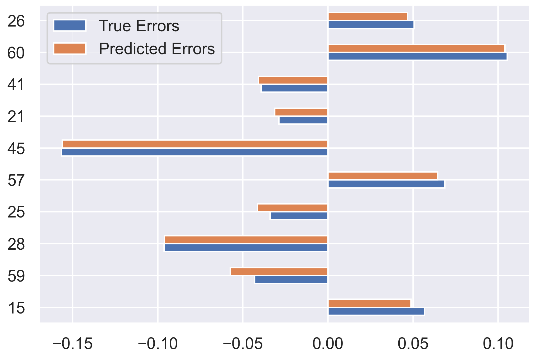}
		\caption{Training Size (90\%)}
	\end{subfigure}
	\hfill
	\begin{subfigure}{0.45\textwidth}
		\includegraphics[width=\textwidth]{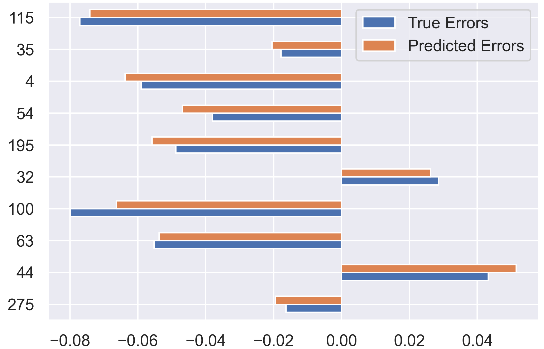}
		\caption{Training Size (55\%)}
	\end{subfigure}
	\hfill
	\caption{Dataset: Experimental dataset}
	\label{fig:22}
\end{figure}

Figures \ref{fig:14} - \ref{fig:22} show the results of the error estimation component of the confidence net on the test data of each dataset. Although the training size has been reduced to almost half of the dataset, the confidence nets model maintains robust error prediction performance on most test datasets. \\
As seen in the results, the confidence nets error estimation component can estimate several weaknesses of the neural network's predictions; in some cases, the error can be reduced to less than 0.05. It is interesting to note that the error estimates are incredibly accurate for the rectangular concrete columns dataset, foamed concrete dataset and Experimental dataset, which have dataset sizes of 623, 1133, and 639, respectively. We deduce that the reason may be due to the high-quality data in these datasets, resulting in reasonable estimations. In other datasets such as (\cite{Yeh1998, AltayebMohamedelmujtaba;WangXin;Musa2021,RamagiriKruthiKiran;BoindalaSrimanPankaj;Zaid2021,JeongHoseong;HanSun-Jin;ChoiSeung-Ho;KimJaeHyun;Kim2020}, the confidence nets error estimation model performance is satisfying, but the accuracy is reduced slightly. On a few occasions, the confidence nets model makes drastic mistakes, and the error is estimated as a positive error when it is a negative error (i.e. the correction makes the prediction worse). However, this is seen rarely and can be due to human error in these data points. For the time being, the confidence nets model is incapable of modelling a probability of human error and does not take any measures to quantify it. The worst performance is seen in the predictions performed on the Concrete Slump Test dataset (\cite{Yeh2007}) and (\cite{IrandeganiMohammadAli;ZhangDaxu;Shadabfar2021}). This is expected as these two datasets have the smallest number of samples; thus, training data is minimal and may not provide proper weight updates to reach an excellent local optimum. This further deteriorates the performance of the test dataset. \\


\subsection{Error Inclusion}
Table \ref{tab:inclusion_results} shows the results of the comparison between ANNs and the Confidence nets model. As expected, ANNs without any modifications performed poorly on small datasets. The only exception is the 622 samples dataset of RCFST columns, where all the predictions were correctly within the average training error. The confidence-nets model provides much more robust and consistent performance on small datasets than the base ANN model. In the case of experimental data, the results are significantly improved compared to the base ANN.

\begin{table}[H]
	\caption{The calculations of error inclusion for two training sets sizes are shown (90\% and 55\%). The higher performance is highlighted in bold font.}
	\centering
	\begin{tabular}{p{0.03\textwidth}p{0.4\textwidth}p{0.1\textwidth}p{0.1\textwidth}p{0.1\textwidth}p{0.1\textwidth}}
		\toprule
		\# & Dataset Name  &  \multicolumn{2}{c}{Training Size (90\%)} &  \multicolumn{2}{c}{Training Size (55\%)}   \\
		\cmidrule(r){3-6}
		. & . & ANN     & Confidence-net & ANN & Confidence-net \\
		\midrule
		
		1 & Concrete Slump Test (\cite{Yeh2007}) & 0\% & \textbf{72\%} & 0\% & \textbf{42.5\%} \\
		2 & Concrete Compressive Strength (\cite{Yeh1998}) & 0\% & \textbf{83\%} & 0\% & \textbf{77\%} \\
		3 & Dataset of 622 Rectangular Concrete-filled steel tube columns (\cite{NguyenThiMai2021}) & \textbf{100\%} & 84\% & \textbf{100\%} & 79\% \\
		4 & Strain Hardening Cementitious Composites Data for FDNN ensemble predictive model (\cite{AltayebMohamedelmujtaba;WangXin;Musa2021}) & 0\% & \textbf{63\%} & 0\% & \textbf{63\%} \\
		5 & Compressive strength dataset of foamed/normal concrete for prediction model (\cite{NGUYENTUAN;NGO2018}) & 0\% & \textbf{83\%} & 0\% & \textbf{79\%} \\
		6 & Compressive strength of alkali-activated binder concrete cured at ambient temperature (\cite{RamagiriKruthiKiran;BoindalaSrimanPankaj;Zaid2021}) & 0\% & \textbf{68\%} & 0\% & \textbf{61\%} \\
		7 & A Dataset of Compressive Strength of FRCM-confined Concrete Columns (\cite{IrandeganiMohammadAli;ZhangDaxu;Shadabfar2021}) & 0\% & \textbf{68\%} & 0\% & \textbf{61\%} \\
		8 & Fake transfer length data generated using generative adversarial networks (\cite{JeongHoseong;HanSun-Jin;ChoiSeung-Ho;KimJaeHyun;Kim2020}) & 0\% & \textbf{75\%} & 0\% & \textbf{74\%} \\
		9 & Experimental dataset (\cite{ChenHuaicheng;QianChunxiang;LiangChengyao;Kang}) & 0\% & \textbf{97\%} & 0\% & \textbf{95\%}\\
		
		\bottomrule
	\end{tabular}
	\label{tab:inclusion_results}
\end{table}

\section{Discussion}
Although the Confidence nets provide higher performance on smaller datasets, it is still in their infant stages of development, and the concept is far from perfect. However, as seen from the results in section 3, the concept can work reliably with datasets of size $\geq$ 241 (the smallest dataset that provided good results). Not only this, but the results of the inclusion test show that although the confidence nets concept is reasonably simple, it is more efficient than traditional neural network models. Furthermore, it provides a more informative prediction and is a step towards constructing a reliable prediction interval. It must be taken into account that the prediction interval of the confident nets is always trying to detect weaknesses of the neural network instead of just providing general estimates. The dataset sizes demonstrated are possible to collect in research, and the confidence nets model has proven that it can make the most of small datasets. The error estimation model provides significant assistance in improving the neural network model predictions, and we expect this to provide more robust performance for more complex neural network architectures. The model is expected to generally increase in size per the properties of the XGBoost model when trained against large datasets. As the decision trees grow with training data, the model will correspondingly increase in size. For this reason, we used the XGBoost algorithm, which may be much more efficient than other tree-based methods. Perhaps, this increase in size is one main disadvantage that must be overcome in the future. However, in the meantime, the Confidence nets model can be used as a solution for conducting experiments where an expected error interval is needed. In its current form, the concept of the confidence nets model can be implemented in fields where conducting experiments is expensive, such as civil engineering, material design, etc...\\
\section{Conclusion}
In this paper, we proposed a simple idea for providing reliable predictions while estimating errors using an ensemble learning technique named confidence net. The confidence nets model is a step towards constructing prediction intervals for regression neural networks. This approach has proven to significantly increase accuracy on small datasets without adding further complexity to the neural network's architecture or affecting the application of transfer learning. Moreover, the constructed Prediction Interval gives an informative indicator of the model's confidence in a prediction. The proposed method is tested on various datasets, and a significant improvement in the performance of the neural network model is seen. \\
\begin{itemize}
	\item The confidence nets model could reliably predict errors even when the training dataset size was decreased from 90\% to 55\%. The performance of the confidence nets largely remained the same. 
	\item The robustness of the model is also tested by increasing the test dataset size up to 45\% of the available data. Still, the model can provide reliable predictions on an average of 83\% of the time.
	\item The model's performance indicates that it is possible to use the current confidence nets model on small datasets of sizes $\geq$241 (the smallest dataset that provided good results). 
	\item Finally, the overview of the estimates shows that the confidence nets model does not tend to provide the largest error margins to include errors. Instead, the model attempts to provide smaller prediction intervals, but this interval is increased by introducing the factors $\omega$ and the dissimilarity factor $d_e$.
\end{itemize}
The above finding shows that the confidence nets ensemble currently applies to many problems and can provide reliable results. However, further studies are needed to confirm the reliability of this concept. We expect that with more developments, the next versions of confidence nets may become more suitable and stable over time, thus providing higher accuracy and reliability for estimating experimental success.

\section{Acknowledgements}
The authors thank everyone involved in the preparation of this preprint.

\bibliographystyle{unsrtnat}
\bibliography{references}  

\end{document}